\documentclass[english]{article}
\usepackage[T1]{fontenc}
\usepackage[latin9]{inputenc}
\usepackage{geometry}
\usepackage{xcolor}
\usepackage{pdfcolmk}
\usepackage{calc}
\usepackage{textcomp}
\usepackage{graphicx}
\usepackage{setspace}
\PassOptionsToPackage{normalem}{ulem}
\usepackage{ulem}
\makeatletter

\newcommand*\LyXHairSpace{\hspace{1pt}}
\providecolor{lyxadded}{rgb}{0,0,1}
\providecolor{lyxdeleted}{rgb}{1,0,0}

\DeclareRobustCommand{\lyxsout}[1]{\ifx\\#1\else\sout{#1}\fi}

\usepackage{mciteplus}

\makeatother

\usepackage{babel}
\begin{document}

\title{Training Optimus Prime, M.D.: Generating Medical Certification Items
by Fine-Tuning OpenAI's gpt2 Transformer Model}

\author{Matthias von Davier\thanks{Address for correspondence: mvondavier@nbme.org: Matthias von Davier,
Ph.D., Distinguished Research Scientist, National Board of Medical
Examiners, Philadelphia, PA, 19105}}

\date{August 21st, 2019}
\maketitle
\begin{abstract}
\textbf{Objective:} Showcasing Artificial Intelligence, in particular
deep neural networks, for language modeling aimed at automated generation
of medical education test items. 

\textbf{Materials and Methods:} OpenAI's gpt2 transformer language
model was retrained using PubMed's open access text mining database.
The retraining was done using toolkits based on tensorflow-gpu available
on GitHub, using a workstation equipped with two GPUs.

\textbf{Results:} In comparison to a study that used character based
recurrent neural networks trained on open access items, the retrained
transformer architecture allows generating higher quality text that
can be used as draft input for medical education assessment material.
In addition, prompted text generation can be used for production of
distractors suitable for multiple choice items used in certification
exams.

\textbf{Discussion:} The current state of neural network based language
models can be used to develop tools in supprt of authoring medical
education exams using retrained models on the basis of corpora consisting
of general medical text collections. 

\textbf{Conclusion:} Future experiments with more recent transformer
models (such as Grover, TransformerXL) using existing medical certification
exam item pools is expected to further improve results and facilitate
the development of assessment materials.
\end{abstract}

\section*{Objective}

The aim of this article is to provide evidence on the current state
of automated item generation (AIG) using deep neural networks (DNNs).
Based on earlier work, a first paper that tackled this issue used
character-based recurrent neural networks \cite{von2018a}, the current
contribution describes an experiment exploring AIG using transformer-based
language models \cite{vaswani2017a}. 

Time flies in the domain of DNNs used for language modeling, indeed:
The day this paper was submitted, on August 13th, 2019, to internal
review, NVIDIA published yet another, larger language model of the
transformer used in this paper. The MegratronLM (apart from taking
a bite out of the pun in this article's title) is currently the largest
language model based on the transformer architecture \cite{nvidia2019a}. 

This latest neural network language model has >8 billions of parameters,
which is incomprehensible compared to the type of neural networks
we used only two decades ago. At that time, in winter semester 1999-2000,
I taught classes about artificial Neural Networks (NNs, e.g. \cite{rosenblatt1958a}).
Back then, Artificial Intelligence (AI) already entered what was referred
to as AI winter, as most network sizes were limited to rather small
architectures unless supercomputers were employed. On smaller machines
that were available to most researchers, only rather limited versions
of these NNs could be trained and used, so successful applications
were rare, even though one of the key contributions that enabled deep
learning and a renaissance of NN-based AI, the Long-Short-Term-Memory
(LSTM) design \cite{hochreiter1997a} was made in those years. In
2017, I started looking into neural networks again because I wanted
to learn how to program Graphical Processing Units (GPUs) for computational
statistics and high performance computing (HPC) used in estimating
psychometric models \cite{von2016a}. This finally led me to write
a paper on using deep neural networks for automated item generation
\cite{von2018a}, a field that has seen many different attempts, but
most were only partially successful, and involved a lot of human preparations,
and ended up more or less being fill-in-the-blanks approaches such
as we see in simple form as MadLibs books for learners. 

While I was able to generate something that resembled human written
personality items, using a public database that contains some 3000,
and while several of the (cherry-picked) generated items sounded and
functioned a lot like those found in personality inventories \cite{goldberg1999a,goldberg2006a},
I was somewhat skeptical whether one would be able to properly train
neural networks for this task, given that it would require a very
large number of items, and I assumed that each network for that purpose
would need to be solely trained on items of the form it is supposed
to generate. Part of my concern was that the items that were generated
had to be hand-picked, as many of the generated character or word
sequences ended up not to be properly formed statements. However,
those that were selected for an empirical comparison with human coded
items were found to show the same dimensionality \cite{von2018a}
and hence to be fully useful as replacements of human authored items.
Nevertheless, some doubt remained due to the needed handpicking and
the limited supply of training material, after all, AI and neural
networks have a long history (e.g., \cite{rosenblatt1958a}; also
\cite{MIT1961a}) and have been hyped to be the next big thing that
may soon replace humans and take our jobs. 

As mentioned, items generated using RNNs \cite{von2018a}, then cherrypicked,
were passing empirical evaluations and hence functioned a lot like
the human written items in an online data collection. However, many
of the generated items were either not properly formed statements
that are typical for this domain, or if the network was trained too
long on too little data, they were almost exact copies of what was
entered as training material. Therefore, I concluded one would need
a lot more data, or an unforseen qualitative jump in deep learning
that I expected to be years away. This was wrong, it turns out that
time indeed flies, and the field of deep learning did not rest, and
while in the paper published in 2018 I stated that operational use
could be years away, I am not so sure anymore that we have to wait
that long.

It may well be that we will see automated item generation based on
deep learning systems as soon as 2021 in tools that support item writers
for developing test questions for high-stakes exams, and that deep
neural networks will be used to generate questions or distractors
for multiple choice questions used in test preparation and practice
exams much sooner. The reason why I believe this has to do with a
graduate student who developed a software tool for programmers based
on a product that was released by OpenAI \cite{radford2018a}. The
software that supposedly makes programmer lives so much better is
called TabNine (e.g. \cite{vincent2019a}) and it provides context-sensitive
(intelligent?) auto-completion based on indexed source code files.
The author of the software estimates that TabNine will save programmers
at least 1 second per minute by suggesting how lines of program code
are completed, or what the most likely next line of code may be, based
on the code that the programmer provides and the software uses to
improve a predictive model.

The title of this article is a reference to two relevant lines of
inquiry. There was an article with the title ``Doctor A.I.'' \cite{choi2017a}
which described a deep learning approach using generative adversarial
networks (GANs) to generate electronic health records (EHRs) that
can pass as plausible EHRs, and the other is the recently ignited
race around language models that use a specific neural network structure
called transformer, which was an obvious trigger for many references
to the sci-fit toys and movies. The remainder of this article is structured
as follows: The next section introduces language models that are based
on approaches that can be e used to generate the probability of a
next word or language token using information about a previously observed
sequence of words. The following section outlines potential areas
of application and shows select examples of how NN-based language
models could be utilized in medical licensure and other assessment
domains for AIG.

\section*{Background and Significance}

AIG has been an area of research in the field of employment and educational
testing for quiet some time \cite{bejar2002a}. Employing human experts
to develop items that can be used in medical licensing and certification
is particularly cost intensive, as expert knowledge is needed to author
case vignettes and to develop plausible response options when writing
multiple choice test questions. Any technology that can reduce these
development costs by applying machine learning or AI would be a welcomed
addition to the toolbox of test developers. AIG often either focused
on items that are language free such as intelligence test with matrices
of graphical symbols that need to be completed by test takers \cite{embretson1999a}
or employed methods that amount to something that bears strong similarities
to fill-in-the-blanks texts such as the ones found in MadLibs. 

The current work builds on and extends a study presented by \cite{von2018a}
in which a RNN was trained on an open access database of 3000 items
available through the IPIP database \cite{goldberg1999a}. While this
previous study concluded that with existing recurrent network based
models, and with limited item banks of limited, a practical use of
AI for AIG would be years away, the development of language models
took a quantum leap when researchers did away with recurrence, and
focused on network architectures built around self-attention \cite{vaswani2017a}.
This allowed designing a simple network structure that was easily
trained, allowed parallelism in training, and could be pre-trained
on general corpora of texts and subsequently trained for specific
purposes. 

Retraining (and re-implementation) of the transformer has led to a
variety of applications, including the generation of poems, patent
texts, and completion of code in support of software developers. These
applications will be references in appropriate sections below. The
current article describes a similar experiment with the goal to provide
a tool for developing medical eductaion test items using deep learning
based language models.

\section*{Materials and Methods}

The basis of these predictive approaches are sequential models that
provide the probability of the next word (or other language token
such as full stop, newline, etc.) given a number of previous words.
These models are not new, I recall my first encounter of this type
of model was an article in 'Scientific American' before 1985 when
I was still a high school student and part-stime programmer working
for a small educational gaming company located in Northern Germany
(yes, game-based learning existed back then). This 1980-version actually
goes back to the seminal paper by \cite{shannon1948a}, and constitutes
a primitive language model. This simple model did of course not have
the many layers and the complex network architecture of deep learning
applications that are nowadays used for machine translations, picture
annotations, or automated item generation \cite{von2018a}, but was
much rather based on a single layer that connected an input word (previous
encounter) to an output word (next encounter). Technically, the basis
of this model was a transition matrix, with input (previous) and output
(next) words coded as binary vectors, and the model basically implemented
the Markov assumption for as a model for natural language.

\subsection*{Markovian Language Models}

The model mentioned above is simple language model that can be viewed
as direct translation of the Markov assumption for modeling a sequence
of words $w_{t}\in\Omega_{V}$ with index $t=1,\dots,T.$ Here, $\Omega_{V}$
is a finite set of words, the vocabulary of a language and $S=\left|\Omega_{V}\right|<\infty$
denotes the size of the vocabulary. Let $\omega:\left\{ 1,\dots,S\right\} \mapsto\Omega_{V}$
be an index, i.e., a bijective function that maps integers to words.
That is we can obtain an integer that represents a word $w_{t}$ by
applying $i_{t}=\omega^{-1}\left(w_{t}\right)$ and the associated
word can be retrieved from any integer $i_{t}\in\left\{ 1,\dots,S\right\} $
through $\omega\left(i_{t}\right)$.

In this most simple case of a language model we assume that 
\[
P\left(w_{t+1}\mid w_{t},\dots,w_{1}\right)\equiv P\left(i_{t+1}\mid i_{t},\dots,i_{1}\right)=P\left(i_{t+1}\mid i_{t}\right)\equiv P\left(w_{t+1}\mid w_{t}\right)
\]
for any $t\in\left\{ 1,\dots,T-1\right\} $ namely that the probability
of observing a next word $w_{t+1}$ at position $t+1$ of the sequence
depends only on the last observed word $w_{t}$, and nothing else.
The whole sequence preceding the next to last word is ignored in this
model. Then, if we assume homogeneity of the transitions, i.e., $P\left(\omega^{-1}\left(w_{t+1}\right)\mid\omega^{-1}\left(w_{t}\right)\right)=P\left(\omega^{-1}\left(w_{u+1}\right)\mid\omega^{-1}\left(w_{u}\right)\right)$
whenever $w_{t}=w_{u}$ and $w_{t+1}=w_{u+1}$. With this we can define
\[
M_{t\mid t-1}=\left(\begin{array}{ccc}
P\left(1\mid1\right) & ... & P\left(S\mid1\right)\\
... & P\left(i\mid i\right) & ...\\
P\left(1\mid S\right) & ... & P\left(S\mid S\right)
\end{array}\right)
\]

which is a transition matrix that provides a conditional probability
distribution for any $i=\omega^{-1}\left(w\right)$. If there are
no constraints, this transition matrix has $S\left[S-1\right]=SS-S$
parameters, i.e., roughly the square of the vocabulary size. The parameters
can be obtained by estimating simple sample statistics, or by some
other methods (reference).

A more complex language model would take into account more than one
previous word. This can be implemented as follows. In order to take
the previous $L$ words into account, define $n_{t}=\left(w_{t},w_{t-1},\dots,w_{t-\left(L-1\right)}\right)\in\bigotimes_{i=1}^{L}\Omega_{V}$,
which is an n-gram of length $L$.

Then assume for $t>L$ that
\[
P\left(w_{t+1}\mid w_{t},\dots,w_{1}\right)=P\left(w_{t+1}\mid n_{t}\right)=P\left(w_{t+1}\mid w_{t},w_{t-1},\dots,w_{t-\left(L-1\right)}\right).
\]
While this is a perfectly sound definition, it has practical implications
that may make applications impossible, as soon as the vocabulary contains
more than a few handful of words and the length of sequence, $L$,
grows larger than, say, 3. The issue is that the mini-sequence $n_{t}$
is an element of $\bigotimes_{i=1}^{L}\Omega_{V}$, a much larger
set, with $S^{L}$ elements. For a vocabulary of only 100 words and
three-word sequences, there are already $100^{3}=1,000,000$ different
elements.

A transition matrix that contains all conditional probabilities for
the next words, given the previous three, we would need to train,
estimate, or otherwise obtain $\left(100-1\right)\times1,000,000=\left(S-1\right)\times S^{L}$
probabilities. Therefore, most traditional approaches to construct
such a large transition matrix have not been pursued, as this would
require very large amounts of data.

\subsection*{Char- and Word- RNNs}

One way of circumventing the need to use classical statistical estimation
methods, and to be able to ignore some of the more rigorous requirements
of these methods, is using NNs for the purpose of language modeling.
NNs have been shown to be universal function approximators (e.g. \cite{hornik1991a,hanin2017a}).
This means that an NN with proper design can be used to plug in an
estimate of a function that is otherwise hard to calculate, or hard
to specify based on more traditional approximation or estimation methods.
This advantage is paid for by having only vague knowlegde about the
actual form of the function that is being approximated, as NNs operate
as black boxes and do not easily reveal how the approximation is achieved.

In order to further reduce demands, one could model the sequence of
characters rather than words, as natural languages often contain several
thousands of words, while alphabetic languages can be expressed using
a much smaller character set. Therefore, an alternative to word-based
language models using neural networks can be implemented as a character
based language model. A few years ago, Google released Tensorflow
\cite{abadi2016a} a powerful software toolbox to design, train, and
sample from neural networks. This triggered implementation of a variety
of deep learning approaches using this new tool, among these a character
based deep recurrent neural network (Char-RNN, e.g., \cite{ozair2016a}),
and more recently, other architectures that will be described below.
Obviously, there are many more tools for deep learning, and the models
released for further analyses and fine-tuning, as done in the current
study, are typically available in more than one framework. Wikipedia
provides a list of neural network oriented tool kits at .

\subsection*{Attention is all you need}

Recent language models introduced the concept of attention, as a structure
that was part of the neural network architecture aimed at keeping
certain concepts more salient. This was initially implemented in addition
to the recurrent structures of deep learning models designed for sequence
to sequence and language modeling. However, a recent article \cite{vaswani2017a}
proposed an alternative, much simpler structure in which the context
and the attention mechanism would replace the sequential structures
of RNNs. The title of this article is mirrored in the subsection title,
and this article led to multiple language models published in short
succession, one of which was recently released by OpenAI and forms
the basis of the retrained / fine-tuned model presented in the current
study.

The authors of \cite{vaswani2017a} describe the new network structure
as consisting only of decoder encoder layers with multi-headed attention,
which provides a distribution of most likely language tokens, given
a context of a certain length (say 1024 words and information about
their position). Psychoanalysts would probably say that transformers
simulate some form of free association, note that this is even called
self-attention in the literature. Interestingly, the attention architecture
used in the transformer-based models is simpler than what what previously
deemed necessary in language models based on recurrent neural networks
such as the one used in \cite{ozair2016a,von2018a}. This simpler
structure allows much faster training as the transformer architecture
allows parallel processing by means of simultaneously using word and
position encoding rather than encoding the text sequentially. The
drawback is that (currently) only limited lengths of text can be encoded,
as the parallel processing makes it necessary to have the sequence
to be encoded (input) as well as the output to be present as a whole
(for example, sentence-by-sentence), rather than word-by-word.

\subsection*{Reincarnations of the Transformers: GPT-2, Transformer-XL, Grover,
now MegatronLM}

The GPT-2 model was trained by a team of researchers at OpenAI \cite{radford2018a}
using four different levels of complexity of the transformer architecture.
In an unprecedented move, OpenAI only released the two smallest models,
which comprise of network weights amounting to 117 Million and 345
Million parameters, respectively. The larger models are not published
due to concerns of malicious use cases, and contain up to 1.4 Billion
(!) parameters. However, this number was recently toppled by NVIDIA,
publishing the MegatronLM model that includes more than 8 billion
parameters, and making the code available on github \cite{nvidia2019a}.
However, the 1.4 billion OpenAI parameter model remains unpublished,
as it says on the OpenAI website:
\begin{quotation}
\textquotedbl{}Due to our concerns about malicious applications of
the technology, we are not releasing the trained model. As an experiment
in responsible disclosure, we are instead releasing a much smaller
model for researchers to experiment with, as well as a technical paper.\textquotedbl{}, 
\end{quotation}
All gpt-2 models were trained on what OpenAI called WebText, which
is a 40GB database of text scraped from the WWW, excluding Wikipedia,
as OpenAI researchers assumed that Wikipedia may be used by secondary
analysts to retrain/fine-tune for specific topics. As the full model
is not available, this means that the actual performance of the GPT-2
Transformer model cannot be verified independently, and other researchers
can only use and modify (retrain) the smaller models. The examples
presented in this paper are based on experiments with the that contains
345M hyper-parameters.

There are several other transformer based language models that are
currently under active development and are being made available to
researchers for fine-tuning and adaptation to different applications.
Among these are the Transformer-XL \cite{dai2019a} and Grover \cite{zellers2019a},
and most recently MegatronLM \cite{news2019a}. While the NVIDIA model
used a corpus called \_WebText\_ that contains 40GB of data and was
modeled after the corpus used by OpenAI, Grover was trained on 46GB
of real news and can be used to either generate, or detect, fake news. 

This ability to both detect and generate is based on the fact that
all of these approaches can be viewed as probabilistic models that
predict a sequence of new words (fake news, a translation, next poem
lines, next syntax line in a software program) based on the previous
sentence(s) or lines of code. More formally, we can calculate the
loss function 
\[
H\left(T,\hat{P}\right)=-\frac{1}{T}\sum_{t=1}^{T}\log\hat{P}\left(w_{t}\mid n_{t-1}\right)
\]

where $\hat{P}\left(w_{t}\mid n_{t-1}\right)$ is the estimated distribution
of word $w_{t}$ given context (history) $n_{t-1}$. This is an estimate
of the cross entropy, or logarithmic entropy \cite{shannon1948a}
of the observed sequence $w_{0},\dots w_{T}$ given some initial context
$n_{0}$. This quantity can be used to evaluate generated sequences
relative to the dsitribution the loss based on true (human-generated)
sequences to help distinguish them. The cross entropy is a measure
of how well predicted (in terms of expected log-likelihood, e.g. \cite{gilula1994a})
an observed sequence is if a certain model $\hat{P}$ is assumed to
hold. This loss function is also used during training or fine-tuning
in order to evaluate how well the network predicts new batches of
data that are submitted to the training algorithm. 

It is worth mentioning that while all of these are variations on a
theme, the transformer architecture for language modeling has shown
great potential in improving over previous designs in terms of performance
on a number of tasks \cite{devlin2018a}. In terms of the use for
generating test questions, Grover \cite{zellers2019a} may prove useful
in future applications as it was designed to produce and detect fake
news by using 46GB worth of data based on actual news scraped from
the internet. Retraining Grover with targeted assessment materials
around a content domain is one of the future directions to take for
applied research into automated item generation using NN-based language
models.

\section*{Results}

The applications of deep learning and recurrent neural networks as
well as convolutional networks range from computer vision and picture
annotation to summarizing, text generation, question answering, and
generating new instances of trained material. In some sense, RNNs
can b ere viewed as the imputation model of deep learning. One example
of medical applications is medGAN \cite{choi2017b} an generative
adversarial network (GAN) that can be trained on a public database
of EHRs and then be used to generate new, synthetic health records.
However, medGAN can also be considered an 'old style' approach just
as the approach I used for generating personality items \cite{von2018a},
as medGAN was not based on a pre-trained network that already includes
a large body of materials in order to give it general capabilities
that would be fine-tuned later.

The latest generation of language models as represented by GPT-2 are
pre-trained based on large amounts of material that is available online.
GPT-2 was trained on 40GB of text collected from the internet, but
excluding WikiPedia as it was considered that some researchers may
want to use this resource to retrain the base GPT-2 model. These types
of language models are considered multitask learners by their creators,
i.e., they claim these models are systems that can be trained to perform
a number of different language related tasks such as summarization,
question answering and translation (e.g. \cite{radford2018a}). This
means that a trained model can be used as the basis for further targeted
improvement, and that the rudimentary capabilities already trained
into the model can be improved by presenting further task specific
material.

\subsection*{AI based AIG trained on Workstations with Gaming GPUs}

While this should not distract from the aim of the article it is important
to know that some considerations have to be made with respect of how
and where calculations will be conducted. Software tools used for
deep learning are free \cite{abadi2016a} and preconfigured servers
and cloud services exist that facilitate the use of these tools. At
the same time, there are significant cost involved, and in particular
researchers who develop new models and approaches may need multiple
times more time and resources compared to standard applications that
are used to analyze data. The dilemma is that while most tools for
training deep learning system are made freely available, these tools
are worthless without powerful computers. And pointing to the cloud
is not helpful, as the cloud is ``just someone else's computer''
(as memes and geek merchandise proves): High performance hardware
and algorithms that employ parallelism are needed to train these kinds
of networks, either in the form of hardware on site, in a data center,
or rented through the cloud. The training of RNNs as well as transformer-based
language models takes many hours of GPU time, which comes at significant
costs if the cloud is used. For recent language models of the type
of GPT-2 large (1.4B parameters), or Grover Mega, or XLNet the estimated
costs was around \$30K - \$245K (XLNet) and \$25K (Grover-Mega) More
details can be found at \cite{peng2019a} as well as in online forums
discussing the training and retraining of these models.

Obviously, cloud computing services come at a cost, and while new
preconfigured systems pop up daily and prices will decrease due to
reduced hardware cost and competition, any more involved project that
requires training specialized systems, or re-training existing large
models, will incur significant costs as well. The model used in the
current paper was pretrained on several TPUs (specialized Google hardware
for tensor computations) for over a week, and retraining as well as
finetuning will take weeks of GPU-time in order to produce a system
that is useful for a specific purpose. Therefore, building or purchasing
a deep learning computer is one of the options that should be carefully
considered as well as the use of cloud computing or on demand GPU
time such as Vast.AI. Nowadays, even modest hardware such as gaming
desktops can be utilized as most of these contain powerful GPUs for
graphical processing, which can be turned into thousands of processing
units through tool kits such as CUDA provided by the makers of these
graphics cards (e.g. \cite{d2012a}).

\begin{figure}
\caption{Server parts from eBay used to provide the official PISA 2015 data
analysis, and now upgraded and re-purposed for Automated Item Generation.
All you need is processor cores, RAM and GPUs, and an eBay auction
sniper. While cloud computing is an option, the experiments reported
here are time consuming and cloud computing is currently available
at an on demand rate of \$0.80/hour (\$0.45 pre-ordered) per GPU.
Retraining took 6 days on two GTX 1080Ti GPUs obtained and installed
in a 2013 T7610 Dell dual Xeon processor workstation.}

\centering{}\includegraphics[width=12cm]{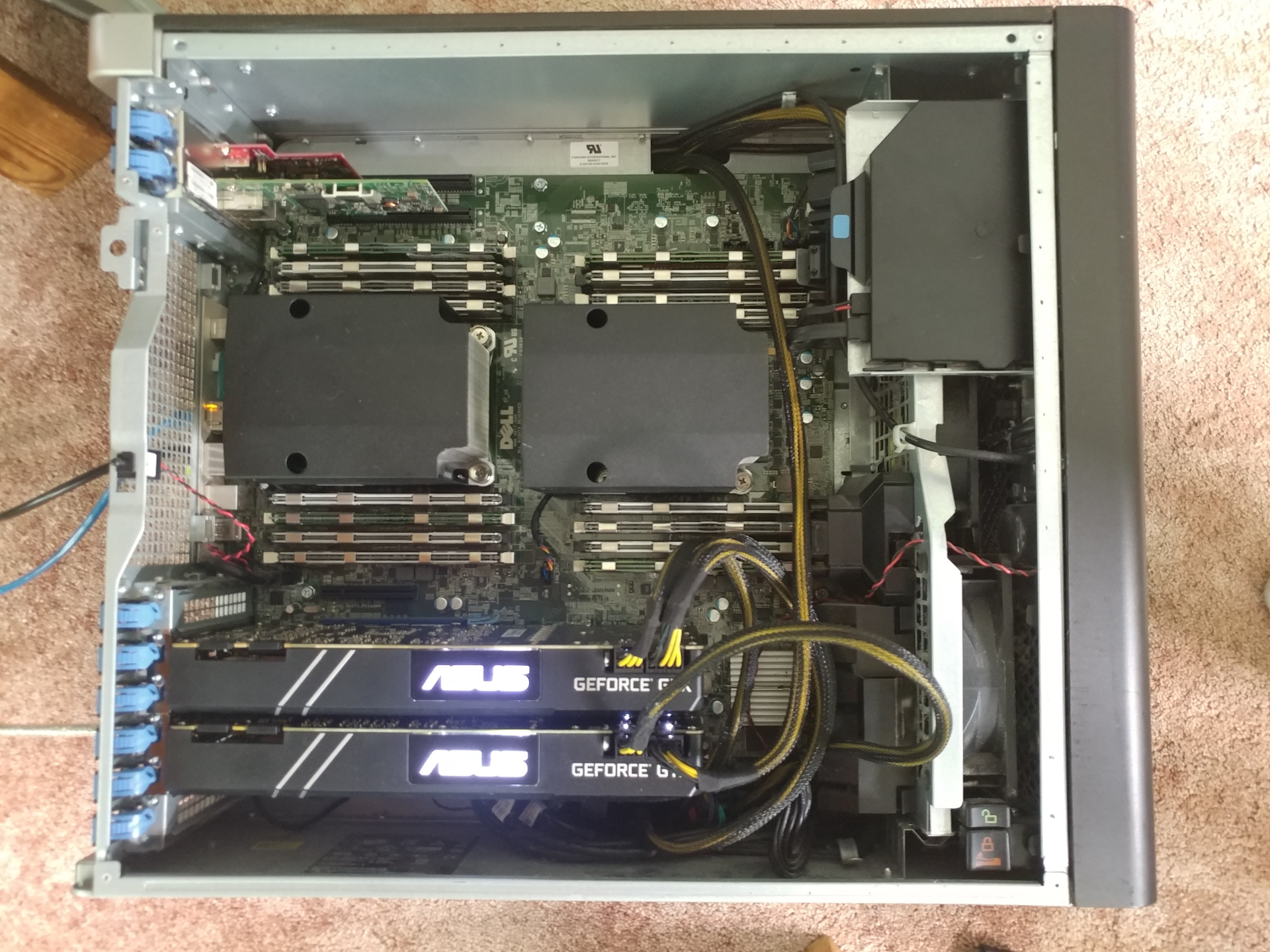}
\end{figure}

The hardware needed for training large NNs can be found at specialized
vendors such as LambdaLabs who often also provide turnkey solutions
such as operating system images that include all the common machine
learning tool-kits such as KERA, TensorFlow, PyTorch and others. An
alternative is to DIY and to use the many web resources that describe
which workstations can be obtained cheaply and how many of the essential
graphical processing units (GPUs) can be housed, with or without modifications.
In addition there are free web resources, for example Google Colab
which is essentially an jupyter notebook that anyone with a google
account can use for deep learning and machine learning experiments
(free for short term use), or time share on demand GPU services such
as Vast.AI can be used for a fee.

Without further digressions, we now turn to how these systems, either
purchased fully configured as turnkey solution, or put together from
used parts, can be utilized to produce text that to a much greater
extent than imaginable only two years ago can facilitate automated
generation of assessment materials, including the generation of electronic
health record, the production of suggestions for distractor choices
in multiple choice items, and the drafting of patient vignettes based
on prompts provided by item writers.

\subsection*{Electronic Health Records and Deep Learning}

The fact that medicine uses IT for storing and managing patient data
brought with it that computer scientists were needed and hired to
work on systems for this purpose. At the same time, data on patients,
as it is stored in electronic health records (EHRs) is highly sensitive,
so that developers working in this area looked for ways to use databases
that would not directly reflect anyone's real data. One way was to
use the same data, carefully anonymized so that individuals cannot
be identified. A second approach was to generate health data of non-existent
patients using the regularities found in real health data.

This was the birth of synthetic EHRs, either in the form of expert-generated
models (Synthea: \cite{walonoski2018a}) or in the form of deep learning
based models that either predict the next clinical event or generate
(plausible) synthetic EHRs based on a training data set of real EHRs
(Dr. AI, \cite{choi2017a}; MedGAN, \cite{choi2017b}). These models
can be used to generate data that can be considered a limitless resource,
and they are anonymous by design, so that concerns about privacy and
data protection are alleviated when using these synthetic records
in design and analysis work. 

A recent systematic review \cite{xiao2018a} describes 98 studies
that use deep learning in conjunction with EHRs. The studies range
from generating synthetic EHRs to enable users to experiment with
these data without any privacy issues to building predictive models
that are used to forecast the next clinical encounter based on existing
EHR history. EHRs are an important source of information, and can
be used to look at systematic differences in trajectories between
patient groups, as well as how different treatments play out by comparing
the prevalence of subsequent encounters.

\subsection*{Distractor Generation through Question Answering}

The utility of the GPT-2 language model is currently explored by means
of retraining for specific purposes. One application that was mentioned
in the introduction is the tabNine software that allows prediction
of line completions and programming code continuation. For that purpose,
each user provides data of their own project related code, which can
further improve prediction.

The ability to generate distractors for existing multiple choice items
is already given even in the original, not retrained 345M GPT-2 model.
The training material contained a large number of cases in which a
question was followed by the prompt A: and an answer. By means of
introducing this type of training material, the model was enabled
to react to a prompt that ends in A: with a continuation of the text
that in many cases represents a meaningful response to a question
that was contained in the prompt. Here, we show two examples that
were generated using a local copy of GPT-2 on the authors workstation
equipped with a GPU and Tensorflow \cite{abadi2016a} which facilitates
sampling from the language model.

The GPT-2 model was trained with material that includes text that
has the structure: ``Q: What is X? A: X is a Y.'', i.e., the model
is prompted to associate a sequence of words that is bracketed in
Q: and A: as a question that requires an answer. The next figure shows
an example of output generated using the 345M model. Note that these
are by far not perfect, but they could serve as inspiration for human
item writers. The first example was generated without any re-training,
using the downloadable version of the 345M gpt2 model.

\begin{figure}
\caption{Example output of the 345M parameter model when used for question
answering:}

\fbox{\begin{minipage}[t]{1\columnwidth - 2\fboxsep - 2\fboxrule}%
\textbf{Model prompt >\textcompwordmark{}>\textcompwordmark{}> Q:
What are the most common side effects of statins? A:}

SAMPLE 1 : They can lead to a reduction in the effectiveness of your
insulin. They can be associated with an increased risk of heart attack
and stroke

SAMPLE 2 : Common side effects are: nausea, vomiting, diarrhea, weight
gain, joint pain. ... 

SAMPLE 3 : Statins can cause stomach inflammation or stomach ulcers.
... 

SAMPLE 4 : There are some common side effects to statins, but these
are not always permanent. For example, people with cardiovascular
disease tend to have more fatigue. 

SAMPLE 5 : These can include: dizziness, feeling like your eyes are
going to pop out (this is called 'repetitive eye movements'), muscle
pain, blurred vision%
\end{minipage}}
\end{figure}

It is clear that not all of the listed side effects are actual ones
patients may experience, however, some overlap with side effects mostly
listed in online resources exists, and some others may be 'plausible
enough' to potentially serve as wrong options in a multiple choice
test. The next example asks about common symptoms of IBS, the selection
of responses were not cherry-picked, and from among two sets of 4
answers, most are on topic.

\begin{figure}
\caption{Responses to a question about symptoms of IBS using a network trained
for 3 days.}

\fbox{\begin{minipage}[t]{1\columnwidth - 2\fboxsep - 2\fboxrule}%
\textbf{Model prompt >\textcompwordmark{}>\textcompwordmark{}> What
are the most common symptoms of irritable bowel syndrome?}

SAMPLE 1.1 In chronic condition, patients report numerous symptoms
including pain, bloating, abdominal pain, diarrhea, constipation,
and vomiting. Those patients who have diarrhea, cough, or abdominal
pain are often affected by this disorder....

SAMPLE 1.2 In 2014, an estimated 3.9 million U.S. adults reported
more than 1 or 2 symptoms of irritable bowel syndrome. About 30 percent
reported at least one symptom. In this survey, 63.8 percent of Americans
...

SAMPLE 1.3 Symptoms of irritable bowel syndrome tend to be more severe
in those with irritable bowel syndrome, which may make them the least
likely to seek diagnosis and treatment. Some symptoms that may be
present include abdominal pain, bloating,  ...

SAMPLE 1.4 Symptoms of irritable bowel syndrome Complications of irritable
bowel syndrome Symptoms of irritable bowel syndrome. Control Medication
Inflammation Abnormal bowel movements. Definitions Irrit ...  

SAMPLE 2.1 Symptoms of irritable bowel syndrome No. of Possible Causes
8.2 10.7 Major Common Symptoms \textendash{} upper and lower gastrointestinal
issues. 1. Irritable bowel syndrome with diarrhea (ILS) has been described
as  ...

SAMPLE 2.2 It is common for the majority of irritable bowel syndrome
patients to experience a variety of symptoms including abdominal pain,
cramping, bloating, nausea, abdominal pain, bloating, diarrhea, cramps,
bloating, cramping  ...

SAMPLE 2.3 My best friend and I recently purchased a new home to be
closer to family and friends, and we felt more comfortable living
at home. However, due to some suspicious and unusual events, we were
contacted by our physician to take our  ...

SAMPLE 2.4 Symptoms of irritable bowel syndrome are often severe and
often require hospitalization. The symptoms of diarrhea (1.2\%) and
constipation (1.4\%) are common in the community. Severe diarrhea
and const ...%
\end{minipage}}
\end{figure}

It is important to note that the responses are based on a general
language model that has not been trained specifically to answer questions
about medical content. This model is, on top of that, the second smallest
of the GPT-2 models, and contains only(?) 345 million parameters,
while other, larger variants contain much more complex model layers
and approximately 1.4 billion parameters \cite{radford2018a}. Again,
note that these responses that could potentially be used as distractor
suggestions were generated without any retraining of specifically
medical assessment materials. 

\subsection*{Automatic Item Generation}

The tests reported in this section are based on the GPT-2 (345M) pre-trained
language model and roughly 800.000 open access subset articles from
the PubMed collection (https://www.ncbi.nlm.nih.gov/pmc/tools/openftlist/)
used for re-training. The data was encoded using the gpt2 (https://github.com/nshepperd/gpt-2)
toolbox for accessing the vocabulary used for pretraining and fine-tuning
GPT-2 using Tensorflow. The 800,000 articles roughly equate to 8GB
worth of text from a variety of scientific journals that allow open
access to some or all of their articles. Training took 6 days on a
Dell T7610 equipped with 128GB RAM, two 10-core Intel Xeon processors,
and two Nvidia 1080 Ti GPUs using CUDA 10.0 and Tensorflow 1.14, Python
3.6.8 and running Ubuntu 18.04 LTS. It was necessary to use the memory
efficient gradient storing \cite{gruslys2016a,chen2016a} options
as the size of data structures for the 345M model used in the retraining
exceeded the 11GB memory of the GPUs without it.

The amount of training data available through open access papers that
can be downloaded from PubMed repositories is quite impressive: The
number of OA articles exceeds 800,000 and the compressed pre-processed
databases used for retraining in this study exceeds 8GB. However,
free medical texts are available in abundance, a 2011 survey \cite{singh2011a}
lists many resources. Language models for data of this size were not
able to be processed on customary hardware only a few years ago, while
nowadays (with a few tricks) even the medium size (345 million hyper
parameter) gpt-2 model can be retrained on decent gaming GPUs.

Incidentally, during the 6 days of training there is some downtime,
which allowed me to find a recent arXiv submission that talks about
automated Patent Application generation using GPT-2 pre-trained with
(public, granted) patent applications available online \cite{lee2019a}.
Other applications include the syntax completion software TabNine
described in the introduction as well as experiments aimed at automatic
generation of poems \cite{gwernnet2019a}. The authors of the GPT-2
patent retraining study used Google Colab, a free online deep learning
platform that allows users access to GPUs for NN training for up to
12 hours. This is insufficient for the 8GB of PubMed data to be fully
retrained on gpt-2 medium, so the author of this paper resorted to
upgrading and using a dual GPU workstation.

The following table shows exemplary results after 2 days of retraining
with the 800,000 article PubMed database. Other publicly available
medical text and article databases are listed in \cite{singh2011a}
. While the results are encouraging, they can certainly not be used
as is, when produced by the NN. However, some editing and human expert
input could use these raw output as inspiration for authoring clinical
vignettes. Results should be closer to human authored item stems using
a transformer that was trained on a large number of real medical licensure
and certification item stems and distractors, and as larger pre-trained
transformer models get published. It should be noted that these early
examples are certainly not texts that would pass as real items. However,
the quality of the texts is such that it can be assumed that the larger
transformer model, trained on real medical licensure items would be
able to produce source material that could be used by medical experts
as prompts to facilitate authoring expert generated items.

\begin{table}
\caption{Example (after only 2 days of training on customary easily obtainable
hardware) of generated text that could potentially be used as the
basis for item stem authoring. Note that the GPT-2 transformer was
retrained on medical open access (PubMed open access subset) articles,
not on any secure or publicly available medical licensure or admission
items. }

{\footnotesize{}Prompt:	A 52-year-old man complaining about chest
pain and headache is coming to the emergency room.}{\footnotesize\par}

{\footnotesize{}PubMed (2 days) retrained, GPT-2 based model-generated
texts:}{\footnotesize\par}

{\footnotesize{}Sample 1:	His blood pressure is 143/99 (normal). He
also has hypertension and urticaria. It is a potentially life-threatening
situation. The first piece of information from the patient is that
he has on occasion been suffering from chronic abdominal pain, abdominal
pain during pregnancy, chronic abdominal pain during childbirth, urinary
tract symptoms. All these are related to chronic abdominal pain. He
is conscious and willing to go to the emergency room for assessment.
However, he is already complaining about abdominal pain when we come
to the hospital.My primary concern is the possibility of heart failure
due to chronic pain. I don't feel that I am in a position to explain
this to the patient, but I want to reassure him that it is a possibility.
So, the decision to take him to the emergency room for evaluation
is already made. Some would argue that he would be better off in the
hospital, but that does not seem to be the case. When it comes to
dealing with chronic pain, we can do better with a diagnosis of dysarthria,
rather than acute cardiomyopathy. We can give him three days of antibiotics.
We can tell him that he will need to have three small intestinal procedures
in the hospital, but that he can go home }{\footnotesize\par}

{\footnotesize{}Sample 2:	He also complained of tingling and aching
in his hands, feet, legs, and arms. The patient is not in good condition.
He had a history of a recent allergic reaction to peanuts. He took
4 mg. of polyphenols: red chip, red, black and white, in combination
with 250 mg. of oat milk and 60 mg. of vitamin B12. He immediately
experienced tingling sensation in his hands and feet. In addition,
he had a mild fever. He did not take medicine for a long time. He
then started to complain of a very severe pain in his chest and abdomen.
His symptoms became worse. He felt cold, tingling sensation, and headache.
He was taken to the hospital, and there a further examination showed
that he was allergic to peanuts. A follow-up examination confirmed
that he was allergic to peanuts. He was discharged home the same day.
A change in his allergy history showed that he had not taken medicine
for a long time, and he had a mild fever. He started to take food
for a long time. He started to take vitamin B12. He started to take
oat milk as well. He started taking 500 mg. of vitamin B12 in combination
with 150 mg. ... }{\footnotesize\par}

{\footnotesize{}SAMPLE 3: The patient was admitted to the hospital
on Friday night with a minor hip fracture. The patient was experiencing
abdominal pain, nausea, and fever. He did not report any cough, other
than the cough of headache and chest pain. The patient was first treated
with antibiotics, then with polyethylene glycol solution (PEG), and
finally with chloramphenicol. At the end of treatment, the patient's
abdominal pain and abdominal pain continued. The patient told his
doctor that he did not feel well, but did not report any cough, cold,
or other symptoms. On the following day, the patient developed a fever
of 40.2\textdegree C and some abdominal pain. On the following morning,
he reported no other symptoms, but fever of 40.2\textdegree C and
fever of 39.2\textdegree C. The patient's temperature was recorded
as 49.3\textdegree C, and his level of red blood cell count was 120/16.
The patient was prescribed phenylalanine (PBA), but did not report
any other symptoms. The patient's BP was recorded as 118/28, his level
of red blood cell count was 126/17. The patient was then referred
to an oncology facility for further treatment. }{\footnotesize\par}

{\footnotesize{}Sample 4:	His symptoms were described as \textquotedbl{}sudden
onset of chest pain with a brief onset of bradycardia. The patient
had a history of transient rapid, transient, and rapid diastolic blood
pressure increases.\textquotedbl{}Patient in critical condition for
dehydration, 5-day interval between study. Recurrent hypertension
and pre-existing coronary heart disease (CHD) with ipsilateral ventricular
tachycardia.Diabetes mellitus. 2 studies (RR\LyXHairSpace =0.75, 95\%
CI\LyXHairSpace =0.65 to 0.85). Over 30-year patient without prior
cardiac disease.Shaolin acupressure technique applied (AG) and exercise
at 1.5m duration.Patient is resting on a reclining couch, \LyXHairSpace \LyXHairSpace a
patient in critical condition for dehydration, 5-day interval between
study; \LyXHairSpace \LyXHairSpace he has chest pain and pain in the
left chest with bradycardia. The patient had no previous CHD, no prior
hypertension, no pre-existing CHD, and a 30-year history of no chronic
obstructive pulmonary disease.Approximate number of patients who participated
in follow-up and follow...}{\footnotesize\par}
\end{table}

A second example contains two sentences as a prompt which provides
a bit more context for the 'attention' ciruits of the transformer
network. The network checkpoint (saved parameters after partial retraining)
that was used represents the retrained GPT-2 after 200,000 cycles
using the 800,000 PubMed open access database.

\begin{table}
\caption{Example 2: Obtained from a network that was retrained for 3 days.
Note that the network recall can be fine-tuned as well to produce
most likely vs. more divergent responses}

{\footnotesize{}Model prompt >\textcompwordmark{}>\textcompwordmark{}>
A 32-year-old woman comes to the doctor's office complaining about
severe headache and blurry vision. She had been suffering from migraine
for years. }{\footnotesize\par}

{\footnotesize{}SAMPLE 1: She was prescribed a drug for tics and headaches,
a sulfate and an antihistamine, and the prescription was confirmed
by the doctor, who thought that the symptoms could be caused by an
active nerve stimulation. It was a decision to keep the dosage, which
was higher than usual. A subsequent results showed a very severe headache,
which did not respond to the medications, and was now very severe.
This was the first case of tetanus-induced brain hemorrhage (HBIH)
in Europe. HBIH is a rare but potentially fatal disease in which the
brain hemorrhage is mainly due to the tetanus toxin.}{\footnotesize\par}

{\footnotesize{}The researchers reviewed the literature to identify
cases of HBIH. All of these cases were reported in the period 2006
to 2011, but a considerable number of cases in the years 2011 to 2012
were only reported in 2005 and 2006. The authors of the paper reported
that four cases of HBIH in Europe have been reported during the years
2006 to 2011, of which three were reported in the period 2006 to 2012.
The second case in this period, which was reported in 2006 to 2011,
was a result of the same kind of drug (antihistamine, sulfate, sulfated
humic acid) as the first case...}{\footnotesize\par}

{\footnotesize{}SAMPLE 2 : At her hospital, she was admitted for medical
examination and he was informed that she had experienced a migraine
and fever in the preceding week. When he examined her eyesight, she
revealed that they were deteriorating. She had been having headaches
for a few years and was very concerned about them. She had the fever
and headache. She was feverish, very fatigued and not able to sleep.
She was in the critical stage and was not able to tolerate even the
medium intensity headaches. She had been very relieved with this problem.
She felt relieved with the headache, but she was in severe pain in
the back of her neck. The doctor who examined her eyesight revealed
that the intensity of her headache was not satisfactory, so the doctors
had recommended that she should be admitted for further medical examination.
She had not been allowed to use a pacemaker since she had been admitted
for treatment of a chronic migraine. However, during her admission,
she complained that her eyesight was getting worse. She was also in
pain in her back. She was admitted for emergency treatment. She was
admitted for treatment of chronic migraine. The doctors who examined
her eyesight gave her an intravenous injection of levodopa and there
were no complications. They also gave her chlordiazepoxide and waited
for a ...}{\footnotesize\par}

{\footnotesize{}SAMPLE 3 : She was wondering if the headgear could
help her to relax and feel a bit better. She was in need of the help
to stop the pain. But she had found that her headache was even worse
than usual. She was planning to take a bath during this period and
it would make her worry. So, she asked the doctor for an eye examination,
to see if there was any abnormality. The doctor, who is a nurse, explained
to the woman that, after a headache, there is an increased production
of nerve gas from the chest and lungs, so it would be best to take
a nasal spray with some corticosteroid treatment. She applied the
nasal spray to her nose and then took a step back and let the cold
air come into the headgear. But, the patient was still not satisfied
and the doctor was still giving her advice. She then asked for another
exam. The patient was now told that she had to take some corticosteroid
treatment. She then was informed that she should take some steroid
nasal spray in addition to the eye examination to take the steroid
treatment. She was then told that her headache was still worse than
usual. On examination, the doctor showed the patient an X-ray of the
neck ...}{\footnotesize\par}

{\footnotesize{}SAMPLE 4: According to her husband, she tried some
anti-emetic medication but there was no difference in her condition.
She felt that migraine pain was like a pressure and pain would only
come when she was moving around. She asked the doctors to prescribe
anti-emetic medication to treat her headache but, she was not satisfied
with the treatment. So, she tried to make a prescription from the
family doctor for anti-emetic medication and the prescription was
filled and passed out the door. When she was with the family doctor,
she tried to find out how to use the anti-emetic medication to relieve
her headache but, the anti-emetic medication was not effective. So,
she finally decided to try the anti-emetic medication and it became
effective. She started to use the anti-emetic medication for about
a month. She finally decided to stop using the anti-emetic medication.
She did not know how to take it and she was very afraid of taking
it. But, she tried to use the anti-emetic medication. And, finally,
the headaches stopped. She was happy and was able to do her normal
activities. Then, she started to take anti-emetic medication and it
was effective. She began to think that she should take...}{\footnotesize\par}
\end{table}

The point to be made here is that the existing network architecture
can be used for question answering, and to a limited extent also for
'inspiration' of human test developers who could enter ideas as prompts
and have the neural network spit out ideas. Current applications that
are similar in kind used the gpt-2 model for retraining based on openly
available patent texts, poems, as well as source code files. It appears
plausible that further fine-tuning with targeted assessment material
should improve the results dramatically, for example by using all
available items in a certain subject domain such as cardiology. It
is not claimed that the current system is fully useful as is, but
the quality of text produced by the currently available transformer
architecture makes it rather likely that correctly formed item stems
can be produced by deep learning based language models in the very
near future.

\section*{Discussion}

We are at the frontier of AI entering many domains of daily life.
While phone makers contribute to the hype and advertise the next generation
of smartphones as running neural networks, there are industrial domains
in which these applications are essential. Among these are computer
vision and assisted driving, others are recommenders for e-commerce,
but also applications that are trained to detect the use of AI for
deep fakes, video material that was made by a machine, programmed
with malicious intent to fool humans. However, there are also many
applications that support human creativity in more benign ways such
as gauGAN \cite{nvidia2019a}, a tool that helps illustrators to compose
landscapes easily with only a few clicks, and tools based on AI that
support wellness \cite{qualcomm2019a} using the same technologies
to analyze data on health that are used to predict what music one
may like based on past purchase and download behavior.

\section*{Conclusion}

The prospects of this technology get really exciting when looking
at how these pre-trained models could be deployed. There are efforts
underway to develop tool-kits that utilize language models, currently
GPT-2 and BERT, another transformer-based language model developed
by Google \cite{devlin2018a}, on iOS devices. This would not train
these networks on phones, but would allow to utilize a trained network
to generate new text based on a context sentence or paragraph entered
by a user. For automated item generation, apps could be developed
that use the language generation on smartphones, for supporting item
developers in writing new content on their mobile devices (https://github.com/huggingface/swift-coreml-transformers).
Once pre-trained models for medical specialties are available, it
would be straightforward to develop a tool in which medical experts
can enter a draft vignette or even a few keywords that are wrapped
by the app into a case description draft, which can then be finalized
and submitted by the human expert for further editing and finalization
by item writers at the testing agency who assembles, administers and
scores the certification tests. At the testing agency, the just developed
case vignette could be finalized using yet another set of machine
learning tools to generate correct and incorrect response options
which are either used in multiple choice formats or for training an
automated scoring system for short constructed responses.

\bibliographystyle{IEEEtranM}
\addcontentsline{toc}{section}{\refname}\bibliography{export3}

\end{document}